\documentclass[journal]{IEEEtran}

\usepackage{cite}
\usepackage{amsmath}
\usepackage{algorithmic}
\usepackage{url}
\usepackage{graphicx}
\usepackage{array}
\usepackage{amssymb}
\usepackage{multirow}
\usepackage[caption=false,font=footnotesize]{subfig}
\usepackage[pagebackref=true,breaklinks=true,colorlinks,bookmarks=false]{hyperref}
\usepackage{amsthm}
\usepackage{dsfont}
\usepackage{booktabs}
\usepackage{xcolor}
\usepackage[T1]{fontenc}
\usepackage{algorithmic}
\usepackage[ruled, vlined]{algorithm2e}
\usepackage{graphicx}
\usepackage{subfig}
\usepackage{booktabs}
\usepackage{threeparttable}

\SetKwInput{KwInput}{Input}             
\SetKwInput{KwOutput}{Output}
\graphicspath{{fig/}}
\def\eg{\emph{e.g.}} 

\def\ie{\emph{i.e.}}

\def\etc{\emph{etc.}} 
\def\vs{\emph{vs.}}
\def\al{\emph{et al. }}

\begin{document}
\title{Learning with Imbalanced Noisy Data by Preventing Bias in Sample Selection}

\author{
	Huafeng~Liu*,
	Mengmeng~Sheng*,
	Zeren~Sun,
	Yazhou~Yao,
    Xian-Sheng Hua,
	and~Heng-Tao Shen
	\thanks{Huafeng~Liu, Mengmeng Sheng, Zeren Sun, and Yazhou Yao are with the School of Computer Science and Engineering, Nanjing University of Science and Technology, Nanjing 210094, China.}
	\thanks{Xian-Sheng Hua is with the Terminus Group, Beijing 100027, China.}
	\thanks{Heng-Tao Shen is with the School of Electronic and Information Engineering, Tongji University, Shanghai 201804, China.}
    \thanks{*Equal Contribution.}
}

\markboth{}%
{Shell \MakeLowercase{\textit{et al.}}: I2CRC}

\maketitle

\begin{abstract}

Learning with noisy labels has gained increasing attention because the inevitable imperfect labels in real-world scenarios can substantially hurt the deep model performance.
Recent studies tend to regard low-loss samples as clean ones and discard high-loss ones to alleviate the negative impact of noisy labels. 
However, real-world datasets contain not only noisy labels but also class imbalance.
The imbalance issue is prone to causing failure in the loss-based sample selection since the under-learning of tail classes also leans to produce high losses.
To this end, we propose a simple yet effective method to address noisy labels in imbalanced datasets.
Specifically, we propose \textbf{C}lass-\textbf{B}alance-based sample \textbf{S}election (\textbf{CBS}) to prevent the tail class samples from being neglected during training. 
We propose \textbf{C}onfidence-based \textbf{S}ample \textbf{A}ugmentation (\textbf{CSA}) for the chosen clean samples to enhance their reliability in the training process.
To exploit selected noisy samples, we resort to prediction history to rectify labels of noisy samples.
Moreover, we introduce the \textbf{A}verage \textbf{C}onfidence \textbf{M}argin (ACM) metric to measure the quality of corrected labels by leveraging the model’s evolving training dynamics, thereby ensuring that low-quality corrected noisy samples are appropriately masked out.
Lastly, consistency regularization is imposed on filtered label-corrected noisy samples to boost model performance.
Comprehensive experimental results on synthetic and real-world datasets demonstrate the effectiveness and superiority of our proposed method, especially in imbalanced scenarios.
The source code has been made available at \url{https://github.com/NUST-Machine-Intelligence-Laboratory/CBS}.

\end{abstract}


%
\IEEEpeerreviewmaketitle

\section{Introduction}
\IEEEPARstart{D}{eep} neural networks (DNNs) have obtained remarkable achievements in various tasks (\eg, image classification \cite{TMM2023_image_classification,Image_classification}, object detection \cite{TMM2023_object_detection,YOLO}, face recognition \cite{Face_Recognition, TMM2023_face_recognition}, instance segmentation \cite{InstanceSegmentation, TMM2023_instance_segmentation,TIP2023_semantic_segmentation,TMM2023_semantic_segmentation2}, natural language processing \cite{NLP}) in recent years.
These successes are highly attributed to large-scale accurately-labeled training datasets (\eg, ImageNet \cite{InageNet}).
Nevertheless, acquiring high-quality manual annotations is expensive and time-consuming, especially for tasks requiring expert knowledge for annotating (\eg, medical images \cite{Medical_Images}).
To obtain large-scale annotated data under a limited budget, recent researchers have started to pay attention to using crowd-sourcing platforms \cite{Crowd-sourcing} or web image search engines \cite{Sesrch_engine} for dataset construction.
Despite reducing the cost of data collection, these methods inevitably introduce low-quality samples that are associated with noisy labels. 
Noisy labels tend to result in inferior model performance due to the strong learning ability of DNNs \cite{zhang2016understanding}.
Therefore, it is significant to develop robust methods for alleviating noisy labels.

\begin{figure}[t]
\centering
\includegraphics[width=\linewidth]{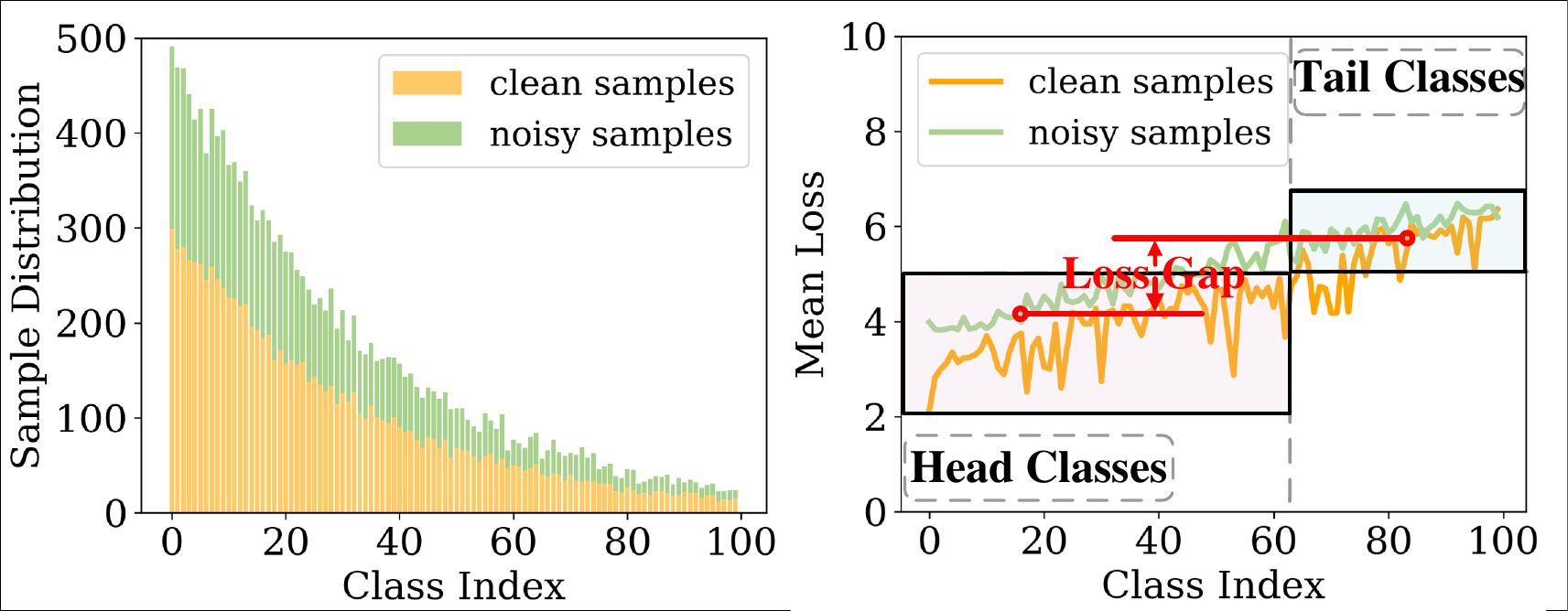}
\vspace{-0.5cm}
\caption{{The sample distribution (left) and the mean loss variation (right) on noisy and imbalanced CIFAR100 (noise rate is 0.4 and imbalance factor is 20). We can find: (1) both tail class samples and noisy samples exhibit large losses; (2) losses of some clean samples belonging to tail classes are even larger than losses of some noisy ones from head classes. 
Accordingly, existing low-loss-based sample selection methods tend to fail when distinguishing clean and noisy samples. This inspires us to develop a class-balanced sample selection method to combat noisy and imbalanced labels.}}
\label{figure1}
\end{figure}

{Recently, some methods have been proposed to address the label noise problem \cite{co-teaching, josrc, JoCoR, Co-teaching+,dividemix,sun2022pnp, SELC, Distribution, LiuTongliang_v1_NIPS, LiuTongLiang_v2_TPAMI, LiuTongLiang_v3_CVPR, LiuTongLiang_v4_CVPR, LiuTongLiang_v5_CVPR, HANBO_v1_TPAMI, HANBO_v2_CVPR}.
Existing approaches mainly employ two kinds of strategies for tackling noisy labels: loss/label correction \cite{M-Correction, PENCIL,dividemix} and sample selection \cite{co-teaching,sun2022boosting, JoCoR}. 
Loss/label correction methods typically attempt to rectify labels by using the noise transition matrix \cite{patrini2017making,goldberger2017,Reviewr2_1} or model predictions \cite{dividemix, josrc, PENCIL}. 
For example, methods such as loss correction \cite{patrini2017making} attempt to first estimate the noise transition matrix and then utilize forward and backward correction to mitigate the impact of label noise.
\cite{Reviewr2_1} proposes a transition-revision (T-Revision) method to effectively learn transition matrices, leading to better classifiers.}
Jo-SRC \cite{josrc} uses the temporally averaged model (\ie, mean-teacher model) to generate reliable pseudo-label distributions for training.
PENCIL \cite{PENCIL} proposes to directly learn label distributions for corrupted samples in an end-to-end manner.
However, loss/label correction methods usually suffer from error accumulation due to the imperfectness and unreliability of the estimated noise transition matrix and model predictions.
Contrarily, sample selection methods primarily seek to divide training samples into a ``noisy'' subset and a ``clean'' subset, and then use the ``clean'' one for training \cite{co-teaching, josrc, sun2021webly, Small_Loss}. 
The effectiveness of recent sample-selection-based approaches is mainly attributed to the \textit{\textbf{Memorization Effect}}: DNNs first fit clean samples and then gradually memorize noisy ones. 
Accordingly, existing methods usually regard samples with small losses as clean ones. 
For example, Co-teaching \cite{co-teaching} cross-updates two networks using small-loss samples selected by its peer networks. 
Jo-SRC \cite{josrc} proposes to employ Jensen-Shannon Divergence for selecting clean samples globally.
DivideMix \cite{dividemix} extracts the clean subset by fitting the loss distribution with the Gaussian Mixture Model.

However, real-world scenarios contain not only noisy labels but also class imbalance \cite{Class-imbalance, InageNet, Clothing1M}. 
Most training data tends to belong to the majority classes (\ie, head classes), while some other classes (\ie, tail classes) may possess only a few training samples.
Class imbalance leans to mislead the optimization of DNNs to sub-optimal solutions, in which models will predict most samples as head classes. 
Samples from tail classes will be under-learned.
Consequently, the generalization performance and robustness of DNNs are inevitably degraded.
Existing approaches designed for noisy labels usually implicitly hypothesize that training samples are class-balanced and thus tend to fail when noisy samples and class imbalance exist simultaneously.
Label correction methods cannot guarantee the reliability of corrected labels since tail classes have far fewer samples than head classes.
Sample selection methods are prone to suffering from learning bias. 
These methods mostly rely on the low-loss criterion. 
Nevertheless, as shown in Fig.~\ref{figure1}, samples from tail classes will also have high losses due to under-learning.
{ Nevertheless, as shown in Fig.~\ref{figure1}, (1) both tail class samples and noisy samples exhibit large losses; (2) losses of some clean samples belonging to tail classes are even larger than losses of some noisy ones from head classes due to under-learning.}

To alleviate the aforementioned issues, we propose a simple yet effective method to learn with noisy labels by balanced sample selection. 
Our method ensures that tail class samples are learned sufficiently by preventing head classes from prevailing in the selected clean samples.
Specifically, we propose \textbf{C}lass-\textbf{B}alance-based sample \textbf{S}election (\textbf{CBS}) to divide training samples into a ``clean'' subset and a ``noisy'' subset in a class-balanced manner.
Subsequently, we propose \textbf{C}onfidence-based \textbf{S}ample \textbf{A}ugmentation (\textbf{CSA}) to minimize the negative effect caused by noisy tail class samples being grouped into the ``clean'' subset. 
By fusing selected clean samples based on confidence, CSA promotes the stability of model training by assuring the reliability of samples fed into the model.
Moreover, to exploit selected noisy samples and avoid the waste of data, we resort to prediction history to rectify labels of noisy samples and feed them into the model afterward.
In order to alleviate the potential harm induced by low-confidence corrected samples (\ie, presumably erroneous corrections), we introduce the \textbf{A}verage \textbf{C}onfidence \textbf{M}argin (ACM) metric to assess the quality of corrected labels.
ACM estimates the contribution of a corrected sample to the model by investigating the gap between its confidence scores of the top-2 candidate corrected labels.
Additionally, ACM leverages the evolving training dynamic, ensuring that low-quality corrected labels are effectively masked out.
Lastly, we design a consistency regularization term to encourage sample-view-wise and epoch-wise prediction consistency, maximizing data exploitation and boosting model performance further. 
Comprehensive experimental results have been provided to verify the effectiveness and superiority of our proposed method.

Our main contributions are summarized as follows :
\begin{itemize}
    \item We propose a simple yet effective approach to address noisy and imbalanced labels. Our proposed class-balanced sample selection assures class balance during the sample selection process to alleviate the learning bias induced by the data imbalance.
    \item We propose to employ confidence-based sample augmentation to enhance the reliability of selected clean samples. The exponential moving average (EMA) is leveraged to correct labels for noisy samples by resorting to prediction history.
    Moreover, consistency regularization is adopted to achieve further model enhancement. 
    \item We propose the average confidence margin metric to measure the quality of corrected labels during training. 
    It quantifies the gap between the confidence scores corresponding to the top-2 candidate corrected labels, thereby ensuring that low-quality corrected noisy samples are appropriately discarded from training.
    
    \item We provide comprehensive experimental results on synthetic and real-world datasets to illustrate the superiority of our approach. Extensive ablation studies are conducted to verify the effectiveness of each proposed component.
\end{itemize}

\section{Related Work}
\subsection{Learning with Noisy Labels}
Label noise in training data has been evidenced to have a detrimental impact on the training of deep neural networks \cite{josrc, sun2020crssc, sun2021webly, ACMMM, TMM2021_web_noisy,ACMMM2020}. 
Existing methods designed for noisy labels can be primarily categorized into the following three groups: label correction \cite{patrini2017making, goldberger2017, PENCIL}, sample selection \cite{co-teaching, JoCoR,josrc}, and other methods \cite{GCE, AGCE, mixup, ELR, Contrastive-Learning-v1, Contrastive-Learning-v2-Sel-CL, TMM2023_fine_grained}. 

{\subsubsection{Label or Loss Correction} 
To cope with label noise, one intuitive idea is to correct sample losses or corrupted labels. 
Methods such as loss correction \cite{patrini2017making} attempt to first estimate the noise transition matrix and then utilize forward and backward correction to mitigate the impact of label noise.
\cite{Reviewr2_1} proposes a transition-revision (T-Revision) method to effectively learn transition matrices, leading to better classifiers.
Goldberger \al \cite{goldberger2017} proposes to use an additional layer to estimate the noise transition matrix.}
Some other researchers focus on correcting labels based on model predictions.
For instance, PENCIL \cite{PENCIL} proposes to learn label distributions according to model predictions. 
Tanaka \al \cite{Label_correction_1} proposes to relabel samples by directly using pseudo-labels in an iterative manner.
However, the noise transition matrix is difficult to estimate accurately, while prediction-based label correction tends to suffer from error accumulation.
Consequently, these methods are prone to struggling with significant performance drops under high noise settings due to the low quality of corrected labels.

\subsubsection{Sample Selection}
Another straightforward idea for addressing noisy labels is to select clean samples and discard selected noisy ones from training. 
For example, Co-teaching \cite{co-teaching} maintains two networks and lets each network select small-loss samples as clean ones for its peer network. 
Co-teaching+ \cite{Co-teaching+} integrates Co-teaching and model disagreement to identify clean samples.
JoCoR \cite{JoCoR} exploits a joint loss to select small-loss samples to encourage agreement between models. 
Besides the popular low-loss-based sample selection, some recent methods propose new selection criteria for finding clean samples.
For instance, PNP \cite{sun2022pnp} simultaneously trains two networks, in which one predicts the category label and the other predicts the noise type.
NCE \cite{NCE} resorts to neighbor data to identify clean and noisy samples.
BARE \cite{BARE} proposes a data-dependent, adaptive sample selection strategy that relies only on batch statistics of a given mini-batch to promote the model robustness against label noise.
Nevertheless, these methods usually rely on the class-balanced hypothesis, rendering them inadequate for addressing noisy and imbalanced datasets in real-world scenarios.
{In this paper, we introduce the class-balance-based sample selection strategy to simultaneously tackle label noise and class imbalance issues. Our method is applied per class, mitigating the loss gap between different classes.}

\begin{figure*}[t]
	\centering
	\includegraphics[width=0.95\linewidth]{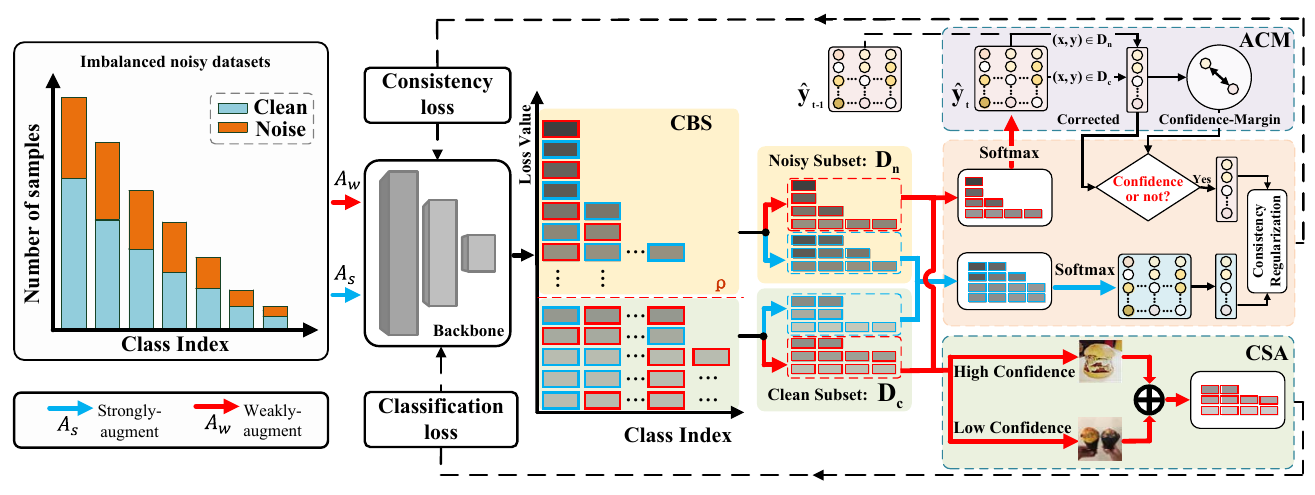}
 \vspace{-0.4cm}
	\caption{{The overall framework of our proposed approach. We first divide the noisy training set into clean and noisy subsets in a class-balanced manner based on the proposed class-balance-based sample selection (CBS) method. Then, for samples in the clean subset, we propose a confidence-based sample augmentation (CSA) method to enhance the reliability of the selected clean samples. Subsequently, the exponential moving average (EMA) is adopted for correcting labels of noisy samples. Thus, noisy samples are also used for model training. Besides, the average confidence margin (ACM) is proposed to measure the quality of corrected labels as the training progresses. Finally, we employ consistency regularization to boost the model performance further. This regularization term can not only enhance the extracted features but also stabilize training by encouraging epoch-wise prediction consistency.}}
\label{fig:pipeline}
\vspace{-0.2cm}
\end{figure*}

\subsubsection{Other Methods}
Apart from the two types of methods mentioned above, there are other attempts that have been established to address noisy labels~\cite{GCE,AGCE,mixup,ELR,Contrastive-Learning-v1,Contrastive-Learning-v2-Sel-CL}.
For example, 
AGCE \cite{AGCE} proposes asymmetric loss functions to address discrete and continuous noisy labels.
ELR \cite{ELR} aims to mitigate the impact of noisy data by applying loss gradient regularization. 
SR \cite{SR} introduces a sparse regularization approach that constrains the network output to a permutation set within a one-hot vector framework.
Recently, some researchers have strived to take advantage of contrastive learning methods.
TCL \cite{TCL} proposes to focus on learning discriminative representations aligned with estimated labels through mixup and contrastive learning.
Sel-CL \cite{Contrastive-Learning-v2-Sel-CL} introduces selective-supervised contrastive learning to learn robust representations and handle noisy labels.

\subsection{Class Imbalance}
Real-world scenarios contain not only noisy labels but also class imbalance, posing a more challenging problem.
Prior works mainly resort to the sample re-weighting strategy for addressing class imbalance \cite{L2RW, Meta-Weight-Net, Curvenet, ULC}. 
These methods usually assign larger weights to tail classes while smaller weights to head classes.
For example, \cite{L2RW} proposes to assign different weights to training samples based on gradient directions. 
\cite{Meta-Weight-Net} proposes a sample weighting function based on meta-learning. 
However, existing approaches are usually vulnerable when training with noisy and imbalanced data. 
It should be noted that noisy and tail class samples exhibit high losses.
Noisy samples require smaller weights, while tail class samples require larger weights.
{CNLCU \cite{Reviewr2_2}, CoDis \cite{Reviewr2_3}, CurveNet \cite{Curvenet} and ULC \cite{ULC} propose initial attempts to address noisy labels and class imbalance simultaneously.
CNLCU \cite{Reviewr2_2} extends time intervals and utilizes the mean of training losses at different training iterations to reduce the uncertainty of small-loss examples.
CoDis \cite{Reviewr2_3} measures the discrepancy by using the distance of prediction probabilities between two networks.
CurveNet~\cite{Curvenet} proposes to learn valuable priors for sample weight assignment based on the loss curves.
ULC~\cite{ULC} performs epistemic uncertainty-aware class-specific noise modeling to identify trustworthy clean samples and refine/discard highly confident true/corrupted labels.}
In this work, instead of following the re-weighting paradigm, we propose a class-balanced sample selection method to ensure that tail classes are sufficiently learned during training.

\section{Methods}
To effectively mitigate the performance degradation caused by noisy labels and class imbalance, we propose to learn from noisy labels by employing balanced sample selection. 
Initially, we partition the training dataset into two subsets (\ie, the clean and noisy subsets) based on our proposed class-balance-based sample selection (CBS) method.
For samples in the clean subset, we further enhance their reliability using the proposed confidence-based sample augmentation (CSA).
For samples inside the noisy subset, we correct their given labels based on the exponential moving average (EMA). 
Besides, we introduce the average confidence margin (ACM) metric to enhance the quality of corrected labels as the training progresses.
Lastly, we incorporate a consistency regularization term to further boost the model performance by encouraging sample-view-wise and epoch-wise prediction consistency.
The overall framework of our approach is shown in Figure~\ref{fig:pipeline}.

\subsection{Preliminaries}
Let $D_{train}=\{(x_i,y_i)|i=1,...,N \}$ be a noisy $C$-class dataset containing $N$ training samples, where $x_i$ denotes the $i$-th image and $y_i\in\{0,1\}^C$ is its associated label (potentially noisy). $y_i^*$ is the ground-truth label of $x_i$. 
We denote $\mathcal{F}(\cdot,\theta)$ as the neural network model parameterized by $\theta$. 
Given an image-label pair $(x, y)$, we optimize the network by employing the loss function $\mathcal{L}(\mathcal{F}(x,\theta),y)$ (\eg, cross-entropy loss) during the training process. 
In the conventional training process, we implicitly assume that the annotated labels of all training samples are accurate (\ie, $y_i=y_i^*$), and use the following cross-entropy loss to optimize the model parameters.
\begin{align}\label{eq_ce}
\mathcal{L}(\mathcal{F}(x,\theta),y)=&\frac{1}{N}\sum_{i=1}^{N}{l}_{ce}(x_i,y_i) \nonumber \\
=&-\frac{1}{N}\sum_{i=1}^{N}\sum_{c=1}^{C}y_i^clog(p^c(x_i,\theta)),
\end{align}
in which $p^c(x_i,\theta)$ denotes the predicted softmax probability of the $i$-th training sample $x_i$ over its $c$-th class.

Due to the existence of label noise, the empirical risk minimization based on the above loss $\mathcal{L}$ leads to an ill-suited solution. 
Recent researchers \cite{NCE, dividemix, UNICON} have attempted to employ the semi-supervised learning (SSL) framework by combining the sample selection and label correction methods. 
First, the sample selection method is adopted to divide the training set $D_{train}$ into a clean subset $D_{c}$ and a noisy subset $D_{n}$. 
Then, the SSL-based method performs label correction on the subset $D_{n}$ and subsequently uses the corrected labels for model training. 
This work also follows the SSL-based paradigm for addressing noisy and imbalanced labels.

\subsection{Class-balance-based Sample Selection}
Due to the memorization effect of deep neural networks, previous studies \cite{co-teaching,JoCoR,dividemix,josrc} usually select low-loss samples as clean ones in each mini-batch $B$.
The conventional low-loss sample selection strategy empirically performs well in class-balanced datasets.
However, these methods tend to suffer from learning bias since the losses of simple class samples tend to be lower than those of hard class samples.
This learning bias issue is amplified when the dataset is not only label-noisy but also class-imbalanced.
As mentioned above, losses of tail classes tend to be higher than those of head classes.
Clean samples belonging to tail classes may even have higher losses than noisy samples from head classes.
The class imbalance substantially prevents current sample selection methods from distinguishing between clean samples and noisy ones.

To this end, we propose a \textit{class-balance-based sample selection} (CBS) strategy to address the concurrent label noise and class imbalance issues. 
Specifically, we first normalize the losses of all samples.
The normalization is done on all samples to map all losses to [0, 1], bringing them into a common scale and making loss comparison more meaningful. 

\begin{align}
l(\mathcal{F}(x,\theta),y)=\frac{l_{ce}-min\{l_{ce}\}}{max\{l_{ce}\}-min\{l_{ce}\}},
\end{align}
in which
\begin{align}
l_{ce}=l_{ce}(\mathcal{F}(x,\theta),y), (x,y)\in D_{train}.
\end{align}

It should be noted that the loss-based sample selection is done per class after normalization, thus mitigating the loss gap between different classes.
Let $D_{sub_i}$ denote the sample set of the $i$-th class. 
{We select the top $\lfloor \rho \frac{|D_{train}|}{C} \rfloor$ small-loss samples from $D_{sub_i}$ as ``clean'' samples belonging to the $i$-th class. 
$\rho$ indicates the sample selection ratio, which is designed to control the number of the selected ``clean'' samples. In practice, we set the value of $\rho$ based on the estimated noise rate $\eta$ (\ie, $\rho = 1- \eta$).}
If $|D_{sub_i}| < \lfloor \rho \frac{|D_{train}|}{C} \rfloor$, all samples in $D_{sub_i}$  will be selected as ``clean'' ones. 
Thus, we can get $D_{c}$ and $D_n$ as follows:\\
\begin{equation}\label{eq:5}
    D_{c} = \mathop{\bigcup}\limits_{i\in \{1,...,C\}}D_{c_i},
\end{equation}
\begin{equation}\label{eq:5_2}
    D_n = D_{train} - D_c,
\end{equation}
in which 
\begin{equation}
    D_{c_i} = \mathop{\arg \min}\limits_{D_{c_i}^\prime \subseteq D_{sub_i}: |D_{c_i}^\prime|=\delta, (x_j,y_j) \in D_{sub_i}} l_{ce}(\mathcal{F}(x_j,\theta),y_j),
\end{equation}
\begin{equation}
    \delta = \min(\lfloor \rho \frac{|D_{train}|}{C} \rfloor, |D_{sub_i}|).
\end{equation}
Our proposed class-balanced-based sample selection prevents samples of tail classes from being neglected in the selection procedure, ensuring their adequate participation in the training process.
Consequently, the network can sufficiently learn from the tail class samples and correctly produce label predictions.

\textbf{Discussion}.
Deep networks usually learn categories that have more samples better than those having fewer ones.
Given that the cross-entropy loss is unbounded, losses of samples from different categories tend to have different scales, resulting in the class-wise loss gap.
This loss gap is prone to hampering sample selection and thereby downgrading model performance.
To address this issue, our method performs sample selection per class using normalized losses.
The loss normalization brings selection metrics to a common scale, making the comparison easier and more meaningful. 
Meanwhile, the class-wise sample selection strategy effectively mitigates the negative impact caused by the loss gap between different categories.

It is also worth noting that although our proposed class-balanced-based sample selection is designed for imbalanced noisy datasets, it is also beneficial for balanced noisy ones. 
In balanced noisy datasets, the learning difficulties of various categories are inconsistent. 
Samples from simple categories tend to yield smaller losses since they are better learned by the network. Contrarily, samples from hard classes usually result in larger losses.
This issue leans to make the trained network have biased and inferior recognition performance. 
By using our proposed class-balanced-based sample selection method, we can alleviate the imbalanced selection results caused by the biased learning ability of the model, thus achieving better model performance.

\subsection{Confidence-based Sample Augmentation}
Resorting to the proposed class-balance-based sample selection strategy, we can effectively distinguish between clean and noisy samples while ensuring that tail class samples are selected sufficiently for the subsequent training.
However, this selection process will inevitably result in some noisy samples from tail classes being misselected into the clean subset.
This issue may lead to a decrease in the model performance.

Therefore, we propose a \textit{confidence-based sample augmentation} (CSA) method for enhancing the reliability of selected clean samples.
To be specific, for each sample $(x_i,y_i) \in D_{c}$ selected by CBS, we randomly choose another sample $(x_j,y_j) \in D_{c}$ and integrate them to obtain ($\widetilde{x_i}, \widetilde{y_i}$) for sample enhancement.
Samples with higher prediction confidence are more likely to be truly clean. 
When integrating the selected two samples, we assign a larger coefficient for the sample whose prediction confidence is higher and a lower coefficient for the sample with lower prediction confidence.
Here, we use the max predicted softmax probability to measure the prediction confidence.
Thus, the generated ($\widetilde{x_i}, \widetilde{y_i}$) is as follows:

\begin{align}\label{eq:6}
\widetilde{x_i}=
\begin{cases} 
l x_i+(1-l) x_j,p(x_i)^{max} \ge p(x_j)^{max}, \\
(1-l) x_i+l x_j,p(x_i)^{max} < p(x_j)^{max},
\end{cases}
\end{align}

\begin{align}\label{eq:7}
\widetilde{y_i}=
\begin{cases} 
l y_i+(1-l) y_j,p(x_i)^{max} \ge p(x_j)^{max}, \\
(1-l) y_i+l y_j,p(x_i)^{max} < p(x_j)^{max}.
\end{cases}
\end{align}
$l = max(l^\prime, 1-l^\prime$), in which $l^\prime$ is sampled from a Beta distribution $B(\Phi,\Phi)$ (In our implementation, $\Phi$ is empirically set to 4). 
$p(x_i)^{max}$ and $p(x_j)^{max}$ denote the max predicted softmax probabilities of $x_i$ and $x_j$, respectively. 

By adopting the proposed confidence-based sample augmentation, we reconstruct the selected clean subset as 
\begin{align}
\widetilde{D_{c}}=\{(\widetilde{x},\widetilde{y}) | (x, y) \in D_{c}\}.
\label{eq:8}
\end{align}
We accordingly enhance the reliability of selected clean samples.
Then, based on Eq.~\eqref{eq_ce}, we calculate the loss on the obtained clean subset $\widetilde{D_{c}}$ as follows:
\begin{align}
\mathcal{L}_{D_{c}} = -\frac{1}{|\widetilde{D_{c}}|}\sum_{(\widetilde{x},\widetilde{y})\in \widetilde{D_{c}}}\widetilde{y}\ log\ p(\widetilde{x},\theta).
\label{eq:9}
\end{align}

\textbf{Discussion}.
It is worth noting that our CSA, inspired by Mixup \cite{mixup}, is designed to minimize the negative effect of selecting noisy samples as ``clean''.
However, unlike Mixup, CSA integrates selected ``clean'' samples from the clean subset and assigns larger coefficients to samples with higher prediction confidence. 
High-confidence samples are more likely to be truly clean.
Accordingly, CSA maximizes data reliability by ensuring augmented data contains at least some clean knowledge, thus promoting generalization performance.
Although this is a rare occurrence, it is still possible that these two samples are both noisy.
When tackling this kind of extreme case where two samples are both noisy, the combination of $y$ (Eq.~\eqref{eq:7}) smooths label distributions and thus slows down the fitting on label noise, thereby effectively enhancing the model’s generalization performance.

\subsection{Label Correction \& Average Confidence Margin}
Discarding selected noisy samples directly leads to a waste of data.
Meanwhile, our proposed sample selection method may introduce another issue: some clean samples belonging to head classes may be mistakenly identified as noisy samples.
Consequently, we follow semi-supervised learning and conduct label correction for selected noisy samples before feeding them to the network.
Moreover, we propose to impose the metric of \textit{average confidence margin} (ACM) to measure the quality of corrected labels by using the model’s training dynamics.
ACM ensures that low-quality corrected samples are appropriately masked out during training.

Considering that the model is inevitable to fit noisy samples in the later stage of training, we resort to the \textit{Exponential Moving Average} (EMA) to achieve more reliable label correction. The corrected labels for noisy samples are formulated as:
\begin{equation}
\hat{y}^t= \alpha \hat{y}^{t-1} + (1-\alpha) p(A_w(x),\theta), (x,y)\in D_{n}.
\label{eq:10}
\end{equation}
$\hat{y}^t$ is the soft corrected label in the $t$-th epoch. 
$\alpha$ is the EMA coefficient. 
$A_w(x)$ represents the weakly augmented view of the sample $x$.
By introducing the prediction history to alleviate the misguidance from erroneously predicted outputs, the correction results are encouraged to be more robust.

Nevertheless, corrected labels with low confidence are not beneficial for model training, as they are essentially akin to noisy labels.
Accordingly, inspired by \cite{MarginMatch, chen2023softmatch, mixmatch}, we introduce the metric of \textit{Confidence Margin} (CM) to measure the quality of corrected labels as follows:
\begin{align}
CM_{j}^t(x) =
\begin{cases} 
\hat{y}_j^t - max_{c \neq j}(\hat{y}_c^t ), j = \arg \max(\hat{y}^t), \\
\hat{y}_j^t  - max(\hat{y}^t), j \neq \arg \max(\hat{y}^t).
\end{cases}
\label{eq:16}
\end{align}
$\hat{y}_j^t$ is the confidence corresponding to the $j$-th class of the soft corrected label $\hat{y}^t$.
$CM_{ \arg \max(\hat{y}^t)}^t$ quantifies the confidence margin between classes with the largest and the second-largest confidence scores in the corrected label distribution.
Consequently, a lower $CM_{\arg \max(\hat{y}^t)}^t$ value indicates greater ambiguity in the model prediction, making the corresponding label correction less reliable.
We also compute $CM^t_{j}$ for the remaining classes $j \neq \arg \max(\hat{y}^t)$, aiming to reflect how these classes confuse the model prediction.

We find that CM only considers the model predictions at the current epoch, making it potentially unstable.
Thus, we further propose \textit{average confidence margin} (ACM)  to average all the margins with respect to the corrected label from the beginning of training until the current epoch $t$ as follows:
\begin{equation}
ACM^t(x) = \frac{1}{t}\sum_{k=1}^{t}CM^k_{\arg \max \hat{y}^t}(x).
\label{eq:17}
\end{equation}
ACM implements an iterative estimation method for assessing the contribution of corrected labels to model learning and generalization during training, providing a more stable measure of confidence for corrected labels.

In practice, we maintain a vector of confidence margins for all classes accumulated during training. 
We dynamically retrieve the accumulated confidence margin of the predicted class (\ie, $\arg \max \hat{y}^t$) at epoch $t$ to obtain $ACM^t$. 
Eq.\ref{eq:16} illustrates that $CM^t_{j}$ is positive for $j = \arg \max(\hat{y}^t)$ and negative for $j \neq \arg \max(\hat{y}^t)$.
Hence, when predictions of the model frequently disagree across different epochs, the confidence margins for $\arg \max \hat{y}^t$ in previous epochs may not consistently be positive, leading to a low $ACM^t$.
In cases where the model predictions show uniformity and stability across epochs in the corrected label, the confidence margins for $\arg \max \hat{y}^t$ in previous epochs are more likely to be positive, resulting in a higher $ACM^t$.
Accordingly, ACM is evidenced to dynamically capture the characteristics of erroneously corrected labels that adversely affect the training process.
We take a linear interpolation of all corrected labels' $ACM$ at $t$-th epoch as a threshold $\mathcal{T}$ to mask out corrected labels with low confidence.
\begin{equation}
\mathcal{T}^{t} = min(ACM)+(max(ACM)-min(ACM))*\tau,
\label{eq:18}
\end{equation}
where $\tau$ is set to control the value of $\mathcal{T}$ (In our implementation, $\tau$ is empirically set to 0.2).

\begin{algorithm}[t]
	\caption{Our proposed algorithm} 
\begin{flushleft}
	\textbf{Input:} The training set $D_{train}$, the test set $D_{test}$, the neural network $\mathcal{F}(\cdot, \theta)$, warm-up epochs $T_{w}$, total epochs $T_{total}$, the sample selection ratio $\rho$, and the batch size $bs$.
\end{flushleft}
	\begin{algorithmic}[1]
		\FOR {$epoch=1,2,\ldots,T_{total}$}
		\IF {$epoch \le T_{w}$}
		\FOR {$iteration=1,2,\ldots$}
		\STATE Fetch $B=\{(x_i, y_i)\}_{1}^{bs}$ from $D_{train}$ 
		\STATE Calculate $\mathcal{L}_{ce} = - \sum_{i=1}^{bs}y_i\mathcal{\log}p(x_i,\theta)$ 
		\STATE Calculate $\mathcal{L}_{cp} = - \sum_{i=1}^{bs}p(x_i,\theta)\mathcal{\log}{p(x_i,\theta)}$
		\STATE Calculate $\mathcal{L} = \mathcal{L}_{ce} + \mathcal{L}_{cp}$
		\STATE Update $\theta$ by optimizing $\mathcal{L}$
            \STATE Obtain ${CM}$ by Eq.~\eqref{eq:16}
            \STATE Obtain ${ACM}$ by Eq.~\eqref{eq:17}
		\ENDFOR
		\ENDIF
		
        \IF {$T_{w} < epoch \le T_{total}$}
        \STATE Obtain $D_c$ and $D_n$ based on Eqs.~\eqref{eq:5} and \eqref{eq:5_2}.
		\FOR {$iteration=1,2,\ldots$}
        \STATE Fetch $B=\{(x_i, y_i)\}_{1}^{bs}$ from $D_{train}$
		\STATE Obtain $\widetilde{{B}_{c}} \subseteq B$ by Eqs.~\eqref{eq:6} and \eqref{eq:7}
		{\STATE Obtain $\hat{y}$ by Eq.~\eqref{eq:10}}
            \STATE Obtain ${CM}$ by Eq.~\eqref{eq:16}
            \STATE Obtain ${ACM}$ by Eq.~\eqref{eq:17}
            \STATE Obtain $\mathcal{T}^t$ by Eq.~\eqref{eq:18}
            \STATE Calculate $\mathcal{L}_{D_c}$ and $\mathcal{L}_{reg}$ using Eqs.~\eqref{eq:9} and \eqref{eq:19}
		\STATE Calculate $\mathcal{L} = \mathcal{L}_{D_c} +  \mathcal{L}_{reg}$
		\STATE Update $\theta$ by optimizing $\mathcal{L}$
		\ENDFOR
		\ENDIF
		\ENDFOR
	\end{algorithmic}
	\label{alg}
\end{algorithm}

After integrating our proposed ACM, we further enhance the model performance through a consistency regularization loss $\mathcal{L}_{reg}$ between the weakly and strongly augmented sample views. 
$\mathcal{L}_{reg}$ ensures that reliable corrected noisy data is effectively utilized as follows:
\begin{equation}
\mathcal{L}_{reg} = - \frac{1}{|D_n^{\prime}|} \sum_{(x,y) \in |D_n^{\prime}|} \hat{y} log\ p(A_s(x),\theta).
\label{eq:19}
\end{equation}
$D_n^{\prime} = \{(x,y) | ACM^t(x) > \mathcal{T}^t, (x,y)\in D_n\}$.
$A_s(x)$ denotes the strongly augmented view of the sample $x$.
$ACM^{t}>\mathcal{T}^t$ is used to mask out corrected labels with low confidence, hindering their induced harm to the model training.
{ It is worth noting that the samples masked out are only a portion of the samples in the noisy subset, whose corrected labels are deemed unreliable.}
By employing this consistency regularization design, we achieve prediction consistency between different sample views, implicitly enhancing the feature extraction of the network.
Furthermore, we also attain epoch-wise prediction consistency for noisy samples, strengthening the stability of the model optimization.
The epoch-wise label consistency is implicitly realized by Eq.~\eqref{eq:10}, which integrates historical and current model prediction results. 
As noted in Eq.~\eqref{eq:10}, $\hat{y}$ contains predictions from previous epochs to alleviate the misguidance from error prediction, thereby enhancing model stability and reliability.


\begin{table*}[t]
	\caption{The average test accuracy (\%) on synthetic CIFAR10 with various noise rates and imbalance factors over the last ten epochs. The best and second-best results are bolded and underlined, respectively. }
	\begin{center}   
\setlength{\tabcolsep}{3.9mm}{
		\begin{tabular}{r|c|ccc|ccc|ccc}
			\toprule
			\textbf{Imbalance Factor} &\multirow{2}{*}{\textbf{Publication}} &\multicolumn{3}{c}{\textbf{1}} \vline & \multicolumn{3}{c}{\textbf{10}}\vline  & \multicolumn{3}{c}{\textbf{50}} \\ \cmidrule(r){3-5} \cmidrule(r){6-8} \cmidrule(r){9-11}
			\textbf{Noise Rate~~~} & & \textbf{0\%} &\textbf{20\%} & \textbf{60\%} & \textbf{0\%} &\textbf{20\%} & \textbf{60\%} & \textbf{0\%} &\textbf{20\%} & \textbf{60\%}\\
			\midrule
			Standard~~~~~~     & -     &91.35&81.18&45.34 &83.58&67.33&30.90 &69.50&52.20&24.00 \\
			Decoupling \cite{Decoupling}     & NeurIPS 2017     &91.00&85.54&69.12 &81.73&75.11&35.25 &70.49&{60.87}&{31.03} \\
                Co-teaching \cite{co-teaching}     & NeurIPS 2018      &91.68&88.82&75.43 &82.94&76.91&32.46 &68.91&55.47&21.34  \\
                Co-teaching+  \cite{Co-teaching+}   & ICML 2019      &91.20&89.04&74.07 &81.65&72.94&24.33 &66.91&48.79&18.87  \\
			JoCoR \cite{JoCoR}  & CVPR 2020       &91.95&89.09&77.19 &83.14&77.17&30.22 &68.24&59.38&19.48  \\
            DivideMix \cite{dividemix}  &  ICLR 2020   &92.96&\underline{91.63}&79.27 &87.43&{79.49} &50.61 &68.51&62.79 &30.88 \\
            CDR \cite{Reviewr2_4}  &  ICLR 2021   &{94.11}&89.02&81.27 &85.55&75.11&47.28 &73.44&59.69&{31.81} \\
			Jo-SRC \cite{josrc} & CVPR 2021       &93.88&90.57&82.47 &\underline{87.79}&76.02&40.20 &\underline{78.10}&60.75&28.67 \\
			Co-LDL \cite{Co-LDL} & TMM 2022      &92.40&90.49&79.14 &82.86&75.83&41.44 &73.77&53.71&25.72 \\
                AGCE \cite{AGCE}   & TPAMI 2023       &92.79&90.09&{82.68} &86.08&78.95&{52.57} &74.12&57.77&29.53  \\
            TCL \cite{TCL}  &  CVPR 2023 &{93.06}&89.47&\underline{85.66} &83.10&82.20&{52.79} &72.35&\underline{66.41}&\underline{35.61} \\
            
            Robust LR \cite{Robust_LR} &  AAAI 2023  &\underline{94.88}&91.06&{84.25}&87.74&\underline{82.44}&\underline{56.94} &73.82&{64.25}&{31.71} \\
			\midrule
            \textbf{Ours~~~~~~}   & -       &\textbf{95.45}&\textbf{94.30}&\textbf{91.79} &\textbf{89.13}&\textbf{86.42}&\textbf{72.49} &\textbf{82.31}&\textbf{75.36}&\textbf{54.13} \\
			\bottomrule
		\end{tabular}}
	\end{center}
	\label{table:3}
\vspace{-0.4cm}
\end{table*}
\begin{table*}[t]
	\caption{The average test accuracy (\%) on synthetic CIFAR100 with various noise rates and imbalance factors over the last ten epochs. The best and second-best results are bolded and underlined, respectively. }
	\begin{center}   
\setlength{\tabcolsep}{3.9mm}{
		\begin{tabular}{r|c|ccc|ccc|ccc}
			\toprule
			\textbf{Imbalance Factor} &\multirow{2}{*}{\textbf{Publication}} &\multicolumn{3}{c}{\textbf{1}} \vline & \multicolumn{3}{c}{\textbf{10}} \vline & \multicolumn{3}{c}{\textbf{50}} \\ \cmidrule(r){3-5} \cmidrule(r){6-8} \cmidrule(r){9-11}
			\textbf{Noise Rate~~~} & & \textbf{0\%} &\textbf{20\%} & \textbf{60\%} & \textbf{0\%} &\textbf{20\%} & \textbf{60\%} & \textbf{0\%} &\textbf{20\%} & \textbf{60\%}\\
			\midrule
			Standard~~~~~~     & -     &68.40&52.91&19.59 &50.82&34.74&11.31 &37.50&23.64&9.42 \\
			Decoupling \cite{Decoupling}     & NeurIPS 2017     &67.88&54.09&22.82 &52.55&39.16&13.88 &40.44&29.60&10.86 \\
                Co-teaching \cite{co-teaching}     & NeurIPS 2018      &67.94&61.33&47.10 &50.75&43.41 &18.13 &38.30&28.44&11.21  \\
                Co-teaching+  \cite{Co-teaching+}   & ICML 2019      &67.57&56.97&35.74 &51.28&38.59&14.24 &39.83&26.64&9.68  \\
			JoCoR \cite{JoCoR}  & CVPR 2020       &68.98&61.63&44.34 &50.87&42.37&20.10 &37.73&28.68&{13.79}  \\
            DivideMix \cite{dividemix}  &  ICLR 2020   &{75.86}&\underline{69.46}&39.38 &\underline{58.33} &48.66&16.12 &44.51&{30.51}&10.26 \\
            CDR \cite{Reviewr2_4}  &  ICLR 2021   &74.86&63.68&42.66 &57.11&41.42&20.10 &42.25&28.38&12.94 \\
			Jo-SRC \cite{josrc} & CVPR 2021       &75.05&67.95&{48.71} &57.12&\underline{50.91}&{23.21} &\underline{46.96}&\underline{37.86}&12.61 \\
			Co-LDL \cite{Co-LDL} & TMM 2022      &71.02&65.01&40.07&46.06&37.24&17.72 &31.54&25.78&12.17 \\
                AGCE \cite{AGCE}   & TPAMI 2023       &72.49&67.07&47.37 &57.37&44.04&22.38 &43.09&31.86&11.91  \\
            TCL \cite{TCL}  &  CVPR 2023   &{74.12}&63.52&\underline{50.20} &56.53&47.36&{24.47} &45.84&29.71&\underline{18.58} \\

            Robust LR \cite{Robust_LR}  &  AAAI 2023  &\underline{76.87}&68.91&48.07&54.44&46.06&\underline{33.34} &39.05&29.52&14.39 \\
			\midrule
            \textbf{Ours~~~~~~}   & -       &\textbf{78.37}&\textbf{75.27}&\textbf{66.57} &\textbf{63.42}&\textbf{56.43}&\textbf{38.67} &\textbf{48.07}&\textbf{42.52}&\textbf{27.30} \\
			\bottomrule
		\end{tabular}}
	\end{center}
	\label{table:4}
\vspace{-0.45cm}
\end{table*}

\subsection{Overall Framework}
The learning procedure of our proposed method is illustrated in Algorithm \ref{alg} and Fig.~\ref{fig:pipeline}.
The final objective loss function in our method is:
{\begin{equation} \label{eq:13}
    \mathcal{L} = \mathcal{L}_{D_c} + \alpha \mathcal{L}_{reg}.
\end{equation}}
$\mathcal{L}_{D_c} $ and $\mathcal{L}_{reg}$ denote the classification loss term and the consistency regularization loss term, respectively.
{ $\alpha$ is the loss weighting factor}

As presented in Algorithm \ref{alg}, similar to existing methods \cite{co-teaching,JoCoR,josrc}, our method starts from a warm-up stage.
Besides the cross-entropy loss $L_{ce}$, we additionally leverage an entropy loss $L_{cp}$ in the warm-up.
By minimizing $L_{ce}$ and $L_{cp}$ during warm-up, we enhance model prediction confidence.
After warm-up, our proposed method first divides the noisy training set into clean and noisy subsets in a class-balanced manner based on the proposed class-balance-based sample selection (CBS) method. 
Then, for samples in the clean subset, we employ confidence-based sample augmentation (CSA) to increase the reliability of selected clean samples.
Subsequently, we correct the labels of noisy samples based on EMA.
We introduce the average confidence margin (ACM) to filter noisy samples whose corrected labels are of low quality by leveraging the model’s evolving training dynamics.
Finally, we impose consistency regularization from two perspectives: 
(1) we encourage sample-view-wise prediction consistency to improve the feature extraction ability; 
(2) we enforce epoch-wise prediction consistency on noisy samples to stabilize the training process. 
The final objective loss integrates the classification loss on clean samples and the consistency regularization loss on noisy samples.

\section{Experiments}
This section focuses on experimental evaluations. 
We first introduce our experimental setup, including datasets, implementation details, evaluation metrics, and baselines.
Afterward, we present experimental results on synthetic datasets (\ie, CIFAR10 and CIFAR100 \cite{CIFAR}) and real-world datasets (\ie, Web-Aircraft, Web-Bird, and Web-Car \cite{webfg}).
These results firmly verify the effectiveness of our method in alleviating noisy labels in class-imbalanced datasets. 
Moreover, we conduct extensive ablation studies to investigate the effectiveness of each component and hyper-parameters in our method. 

\begin{figure}[t]
	\centering
	\includegraphics[width=0.95\linewidth]{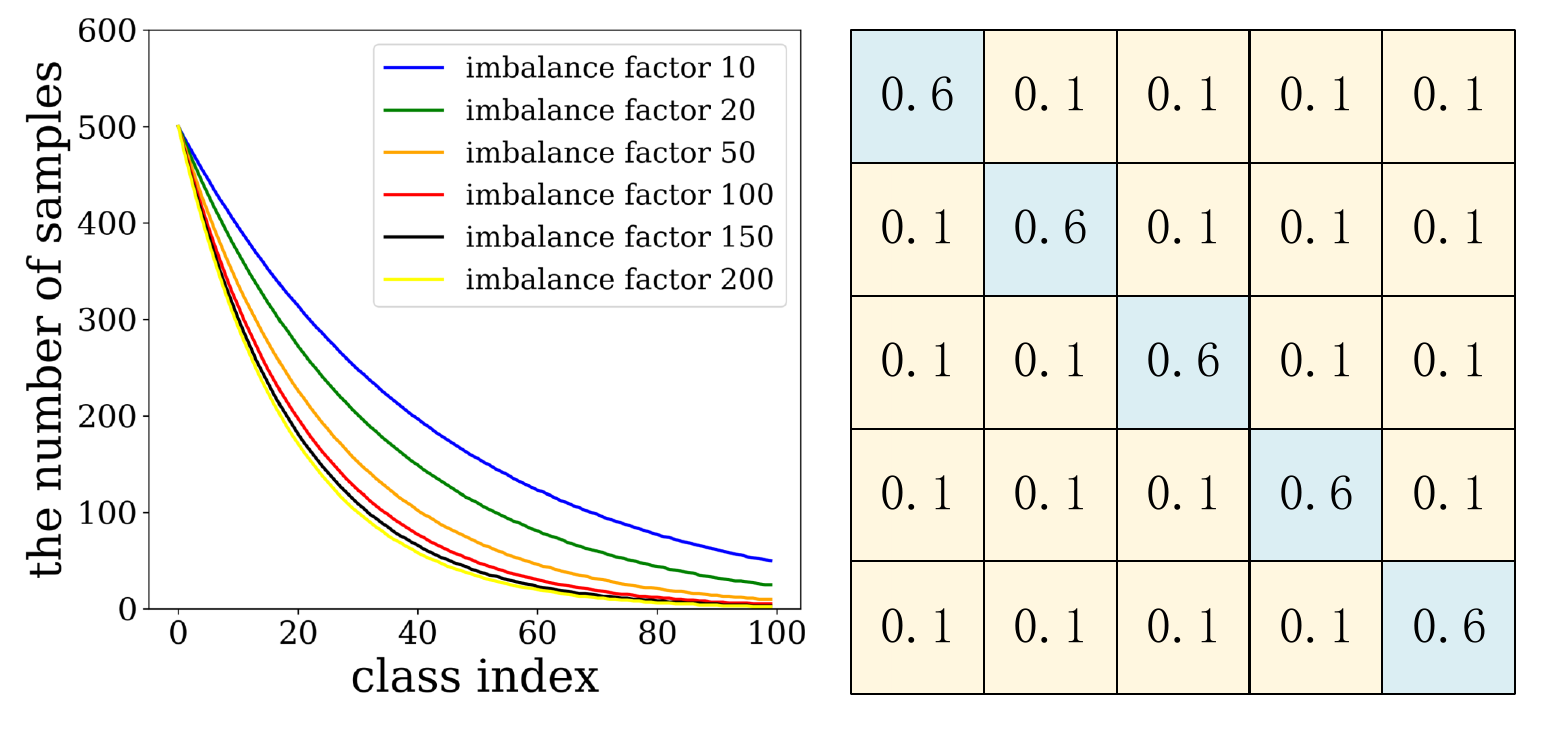}
    \vspace{-0.3cm}
	\caption{The number of samples belonging to each class in CIFAR100 under various imbalance factor settings (left) and an example of the uniform noise transition matrix (right).}
	\label{figure3}
 \vspace{-0.4cm}
\end{figure}

\begin{table*}[t]
\caption{Comparison with SOTA approaches in test accuracy (\%) on real-world noisy datasets: Web-Aircraft, Web-Bird, and Web-Car. The best and second-best results are bolded and underlined, respectively. }
\begin{center}
\setlength{\tabcolsep}{5.2mm}{
\begin{tabular}{r|c|c|c|c|c|c}
\toprule
\multirow{2}{*}{\textbf{Methods~~~~~}}  & \multirow{2}{*}{\textbf{Publication}} & \multirow{2}{*}{\textbf{Backbone}}  & 
\multicolumn{4}{c}{\textbf{Performances(\%)}} \\  
\cmidrule(lr){4-7} & & & \textbf{Web-Aircraft} 	& \textbf{Web-Bird}    & \textbf{Web-Car}  & \textbf{Average} \\ 
\midrule
Standard~~~~~~ & - & ResNet50 & 60.80  & 64.40 & 60.60 &61.93\\
Decoupling \cite{Decoupling} & NeurIPS 2017 & ResNet50 & 75.91 & 71.61 & 79.41 & 75.64\\
Co-teaching \cite{co-teaching} & NeurIPS 2018 & ResNet50 & 79.54 & 76.68 & 84.95 & 80.39\\
Co-teaching+ \cite{Co-teaching+} & ICML 2019 & ResNet50 & 74.80 & 70.12 & 76.77 & 73.90\\
PENCIL \cite{PENCIL} & CVPR 2019 & ResNet50 & 78.82 & 75.09 & 81.68 & 78.53\\
Hendrycks \al \cite{ss-ood} & NeurIPS 2019 & ResNet50 & 73.24 & 70.03 & 73.81 & 72.36\\
mCT-S2R \cite{mCT-S2R} & WACV 2020 & ResNet50 & 79.33 & 77.67 & 82.92 & 79.97\\
JoCoR \cite{JoCoR} & CVPR 2020 & ResNet50 & 80.11 & 79.19 & 85.10 & 81.47\\
AFM \cite{AFM} & ECCV 2020 & ResNet50 & 81.04 & 76.35 & 83.48 & 80.29\\
DivideMix \cite{dividemix} & ICLR 2020 & ResNet50 & 82.48 & 74.40 & 84.27 & 80.38 \\
Self-adaptive \cite{Self-adaptive} & NeurIPS 2020 & ResNet50 & 77.92 & 78.49 & 78.19 & 78.20\\
Peer-learning \cite{Peer-learning} & ICCV 2021 & ResNet50 & 78.64 & 75.37 & 82.48  & 78.83\\
Co-LDL \cite{Co-LDL} & TMM 2022 & ResNet50 & 81.97 & 80.11 & \underline{86.95} & 83.01\\
NCE  \cite{NCE} & ECCV 2022 & ResNet50 & {84.94} & \underline{80.22} & 86.38 & \underline{83.85} \\
SOP  \cite{SOP} & ICML 2022 & ResNet50 & 84.06 & 79.40 & 85.71 & 83.06 \\
SPRL \cite{SPRL} & PR 2023 & ResNet50 & 84.40 & 76.36 & 86.84 & 82.53 \\
AGCE \cite{AGCE} & TPAMI 2023 & ResNet50 & 84.22 & 75.60 & 85.16  & 81.66 \\
TCL \cite{TCL} & CVPR 2023 & ResNet50 & 84.51 & 79.22 & 85.13  & 82.95 \\
Robust LR \cite{Robust_LR} & CVPR 2023 & ResNet50 & \underline{85.78} & 78.65 & 86.13  & 83.52 \\
\midrule
\textbf{Ours~~~~~~~}  & - & ResNet50 & \textbf{87.46} & \textbf{80.50} &	\textbf{87.96} &\textbf{85.31} \\
\bottomrule
\end{tabular}}
\end{center}
\label{table:2}
\end{table*}
\begin{figure*}[t]
	\centering
	\includegraphics[width=0.95\linewidth]{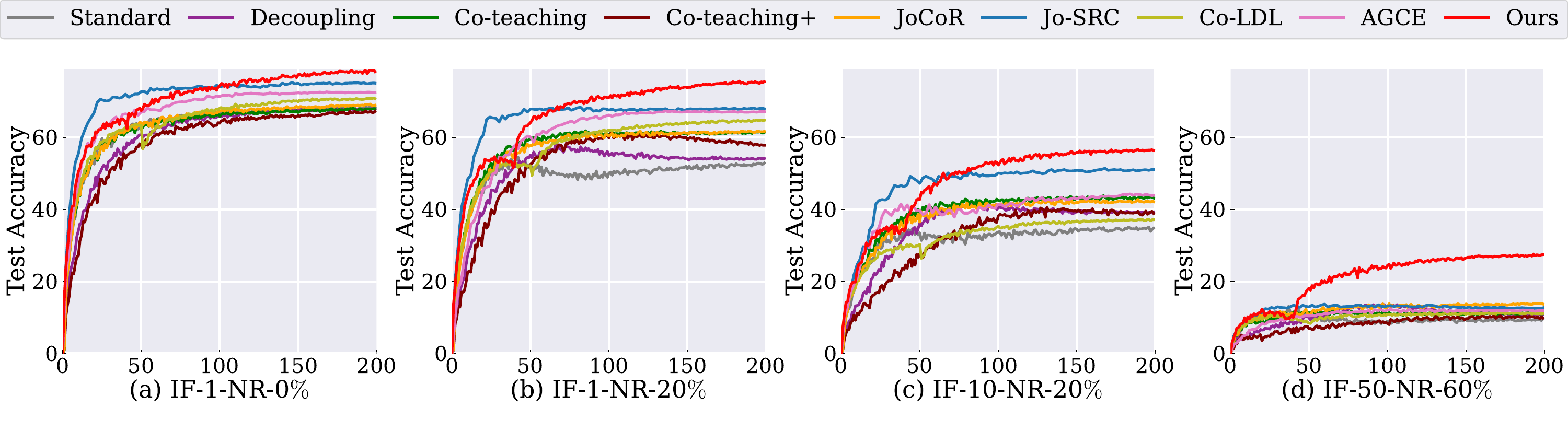}
    \vspace{-0.5cm}
	\caption{ The test accuracy (\%) \vs epochs on CIFAR100 with IF-1-NR-0\% (a), IF-1-NR-20\% (b), IF-10-NR-20\% (c) and IF-50-NR-60\% (d) during the training process. (IF-X-NR-Y\% means that the imbalance factor and the noise rate are X and Y\%, respectively.)}
	\label{figure4}
 \vspace{-0.3cm}
\end{figure*}

\subsection{Experimental Setup}
\textbf{Synthetic Datasets: }
Synthetic datasets are mainly derived from CIFAR10 and CIFAR100~\cite{CIFAR}. 
These two datasets consist of 60,000 RGB images (50,000 for training and 10,000 for testing). 
Images are equally distributed among 10 categories and 100 categories. 
We randomly corrupt the sample labels from their ground-truth categories to other categories using a pre-defined noise rate (NR) $\eta$. 
In our experiments, we adopt the uniform noise, which randomly corrupts labels from their ground-truth classes to other ones with the pre-defined noise rate $\eta$ on CIFAR10 and CIFAR100.
To construct class-imbalanced datasets, we take an exponential function $n_i=n_0\mu^i$ to reduce the number of samples per category, where $n_i$ is the sample number of class $i$ and $\mu \in (0,1]$. 
We use the class imbalance factor (IF), defined as $\frac{max(n_i)}{min(n_i)}$, to measure how imbalanced a dataset is.
Fig.~\ref{figure3} (left) presents the sample distribution of synthetic CIFAR100 under different imbalance factors (\ie, 10, 20, 50, \etc).
Fig.~\ref{figure3} (right) shows an example of the uniform noise transition matrix. 

\textbf{Real-world Datasets:} 
{ To further verify the effectiveness of our method in practical scenarios, we conduct experiments on real-world noisy datasets: Web-Aircraft, Web-Bird, Web-Car \cite{webfg} and Clothing1M \cite{Clothing1M}.}
{These three Web- datasets are subsets of the web-image-based fine-grained image dataset WebFG-496 \cite{webfg}.}
Their training images are crawled from web image search engines, making label noise inevitable.
Web-Aircraft is a fine-grained aircraft dataset containing 13,503 training images and 3,333 test images belonging to 100 different aircraft models. 
Web-Bird is a fine-grained bird dataset containing 200 different classes.
There are 18,388 noisy training instances, whereas the test set consists of 5794 accurately-labeled samples.
Web-Car is a fine-grained car dataset composed of 21,448 samples, and the test set consists of 8,041 samples belonging to 196 car classes.
{ {Clothin1M} comprises 1M clothing images of 14 categories, which are collected from several online shopping websites and include many mislabelled samples.}
These real-world datasets contain both corrupted labels and class imbalance.


\textbf{Implementation Details:} 
On synthetic datasets, we conduct experiments on CIFAR10 and CIFAR100 with various imbalance factors and noise rates. 
We use ResNet18 as our backbone. 
The network is trained using SGD with a momentum of 0.9 for 200 epochs. 
Our warm-up stage lasts for 40 epochs. 
The initial learning rate and batch size are 0.01 and 128, respectively. 
During the robust learning stage, we decay the learning rate in a cosine annealing manner. 
We set $\rho$ and $\tau$ as $1-\eta$ and 0.2, respectively.
{ On real-world datasets (\ie, Web-Aircraft, Web-Bird, Web-Car, and Clothing1M), we follow \cite{Co-LDL} and select ResNet50 pre-trained on ImageNet as our backbone and SGD with a momentum of 0.9 as the optimizer.}
We set the initial learning rate as 0.001, and adopt the cosine schedule to adjust the learning rate during training.
The training lasts for 100 epochs (including 10 warm-up epochs).
{We use random cropping and horizontal flipping as weak augmentation, and adopt AutoAugment \cite{Autoaugment} as strong augmentation.
AutoAugment designs a search space where a policy consists of many sub-policies (\ie, translation, rotation, or shearing), one of which is randomly chosen for each image in each mini-batch.
}

\textbf{Evaluation Metrics:}
We adopt test accuracy as the evaluation metric.
On synthetic CIFAR10 and CIFAR100 datasets, we additionally employ the average test accuracy over the last epochs to evaluate the performance more comprehensively.

\textbf{Baselines:}
We compare our method with state-of-the-art (SOTA) methods for evaluation.
{ On synthetic CIFAR10 and CIFAR100, we compare our method with Decoupling \cite{Decoupling}, Co-teaching \cite{co-teaching}, Co-teaching+ \cite{Co-teaching+}, JoCoR \cite{JoCoR}, DivideMix \cite{dividemix}, Jo-SRC, CDR \cite{Reviewr2_4}, \cite{josrc}, Co-LDL \cite{Co-LDL}, AGCE \cite{AGCE}, TCL \cite{TCL}, and Robust LR \cite{Robust_LR}.}
{On Web-Aircraft, Web-Bird, and Web-Car, the following SOTA methods are adopted for comparison: Decoupling \cite{Decoupling}, Co-teaching \cite{co-teaching}, PENCIL \cite{PENCIL}, Hendrycks \al \cite{ss-ood}, mCT-S2R \cite{mCT-S2R}, JoCoR \cite{JoCoR}, AFM \cite{AFM}, DivideMix \cite{dividemix}, Self-adaptive \cite{Self-adaptive}, Peer-learning \cite{Peer-learning}, Co-LDL \cite{Co-LDL}, NCE \cite{NCE}, SOP \cite{SOP}, SPRL \cite{SPRL}, AGCE \cite{AGCE}, TCL \cite{TCL}, and Robust LR \cite{Robust_LR}).
On Clothing1M, We compare the following methods: Decoupling \cite{Decoupling}, Co-teaching \cite{co-teaching}, JoCoR \cite{JoCoR}, DivideMix \cite{dividemix}, JNPL \cite{JNPL}, UNICON \cite{UNICON} and TCL \cite{TCL}).
}
Moreover, we perform conventional training using the entire noisy dataset. 
The result is provided as a baseline (denoted as ``Standard'').

\begin{table}[t]

\renewcommand\tabcolsep{15pt}
	\caption{Performance comparison with SOTA methods in test accuracy (\%) on Clothing1M.}
	\begin{center}
		\centering
		\begin{tabular}{c|c|c}
			\toprule
			Method    & Publications      & Performance       	\\
            \midrule
            Standard              & -                  & 68.94	\\
            Decoupling \cite{Decoupling}    & NeurIPS 2017           & 69.84	\\
            Co-teaching \cite{co-teaching}    & NeurIPS 2018        & 69.21\\
            JoCoR   \cite{JoCoR}        & CVPR 2020      & 70.30	\\
            DivideMix   \cite{dividemix}    & ICLR 2020             & 74.76	\\
            JNPL \cite{JNPL} & CVPR 2021 &74.15\\
            UNICON \cite{UNICON} & CVPR 2022 & 74.98\\
        
            TCL \cite{TCL}& CVPR 2023              & 74.80 \\
			\midrule
            Ours      &-        & \textbf{74.99}	\\ 
			\bottomrule
		\end{tabular}
	\end{center}
	\label{tab_clothing1m}
 \vspace{-0.4cm}
\end{table}


\begin{table*}[t]
\centering
\setlength{\tabcolsep}{6.5mm}{
\caption{\label{table:7}Effects of different ingredients in test accuracy (\%) on CIFAR10 and CIFAR100 (20\%-10 means that noise rate and imbalance factor are 20\% and 10, respectively). Results at the best epochs are presented.}
\begin{tabular}{l|c|c|c|c|c|c}
\toprule
\multirow{2}{*}{\textbf{Model}}  & 
\multicolumn{3}{c}{\textbf{CIFAR10}} \vline &   
\multicolumn{3}{c}{\textbf{CIFAR100}} \\
\cmidrule(lr){2-7}  &  20\%-1 &20\%-10 	& 20\%-50 & 0\%-10   & 20\%-10  & 60\%-10 \\ 
\midrule
Standard &81.18& 67.33 & 52.20  &50.82& 34.74 & 11.31 \\
\midrule
Standard+CBS &90.95&75.31&60.71 &54.84&42.51&22.47\\
Standard+CBS+CSA &91.11&82.58&68.94 &56.96&50.60&34.72 \\
Standard+CBS+CSA+CR &94.30&85.46&71.87 &62.15&54.46&38.67\\
Standard+CBS+CSA+CR+ACM &94.82&86.42&75.36 &63.42&56.43&39.70\\
\bottomrule
\end{tabular}}
\vspace{-0.3cm}
\end{table*}

\subsection{Experimental Results on Synthetic Datasets}
For evaluating the performance of our proposed method in learning with imbalanced noisy data, we conduct extensive experiments on the synthetic CIFAR10 and CIFAR100 using different imbalance factors (\ie, 1, 10, 50) and noise rates (\ie, 0\%, 20\%, 60\%). 
We compare our proposed method with existing SOTA methods, which are re-implemented using their open-sourced code and default hyper-parameters. 
Results are shown in Table~\ref{table:3}, Table~\ref{table:4} and Fig.\ref{figure4}.

Results presented in Table~\ref{table:3} and Table~\ref{table:4} illustrate:
(1) When the dataset contains only noisy labels (\ie, the imbalance factor is 1), both existing SOTA methods and our method achieve robust performance. Notably, our method achieves the best performance.
(2) When the dataset is noisy and imbalanced, SOTA methods for learning with noisy labels dreadfully degrade their performance when we increase the imbalance factor and the noise rate. 
In particular, methods employing the small-loss criterion (\eg, Co-teaching, Co-teaching+, and JoCoR) generally exhibit inferior performance. 
These methods erroneously discard samples from tail classes due to their large loss values, thereby introducing a learning bias against tail classes.

Our method consistently sustains robust performance, surpassing all competing SOTA methods across all experimental settings.
Results from Table~\ref{table:3} and Table~\ref{table:4} clearly illustrate the effectiveness and superiority of our proposed method. 
This is mainly attributed to our proposed class-balanced sample selection approach.
CBS ensures that the tail class samples sufficiently participate in the model training.
Fig.~\ref{figure4} presents the test accuracy \vs epochs in four different scenarios on CIFAR100.
It further demonstrates that our proposed method consistently performs better than competing approaches during the training process.
Especially in the most challenging situation (\ie, IF-50-NR-60\%), our method still performs superiorly throughout the training process.
These advances in performance validate the superiority of our method.

\begin{figure}[t]
\centering
\includegraphics[width=0.95\linewidth]{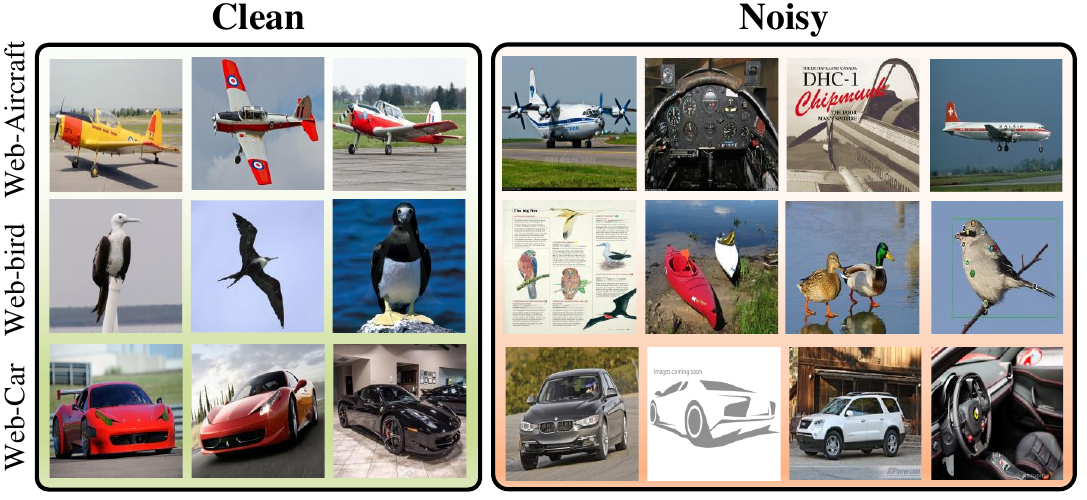}
\vspace{-0.2cm}
\caption{{Some visualization results of clean and noisy samples selected by our sample selection methods on Web-Aircraft, Web-Bird, and Web-Car. The corresponding fine-grained class names are \textit{DHC-1}, \textit{frigatebird}, and \textit{Ferrari 458 Italia Coupe 2012}.}}
\label{figure6}
\vspace{-0.3cm}
\end{figure}
\subsection{Experimental Results on Real-World Datasets}
In addition to evaluating our method on synthetic noisy and imbalanced datasets (\ie, CIFAR10 and CIFAR100), we also utilize three real-world web-image-based datasets (\ie, Web-Aircraft, Web-Bird, and Web-Car) to corroborate the effectiveness and superiority of our method.
Table~\ref{table:2} presents the performance comparison on real-world datasets.
These datasets contain at least 25\% of unknown noisy labels and do not provide any label verification information, making them practical and challenging label noise scenarios.
It should be emphasized that these three datasets are fine-grained ones, which undoubtedly makes them more challenging.
Results of existing methods shown in Table~\ref{table:2} are obtained under the same experimental settings.
As shown in Table~\ref{table:2}, our method consistently outperforms SOTA methods on the three datasets.
{Our method achieves 87.46\%, 80.50\%, and 87.96\% accuracy on test sets of Web-Aircraft, Web-Bird, and Web-Car, respectively.
It achieves a significant performance boost of +1.68\% / +0.28\% / +1.01\% over the best SOTA method.}
In the average test accuracy of these three datasets, our proposed approach outperforms Co-LDL \cite{Co-LDL} and NCE \cite{NCE} by 2.30\% and 1.46\%, respectively.
{
Table~\ref{tab_clothing1m} shows the additional experimental results on another real-world noisy and imbalanced dataset Clothing1M.
By comparing with SOTA methods in Table~\ref{tab_clothing1m}, we can find that our method can achieve competitive performance against SOTA approaches. 
It should be noted that DivideMix, UNICON and TCL involve two simultaneously trained networks, while our method trains only one network.
}

{In order to further visualize the performance of our method, we provide qualitative analysis on three real-world datasets. As shown in Fig.~\ref{figure6}, we provide several visualization results of clean and noisy samples selected by our sample selection methods on three fine-grained categories (\ie, \textit{DHC-1}, \textit{frigatebird}, and \textit{Ferrari 458 Italia Coupe 2012}).
It is evident that our proposed selection method can effectively distinguish clean and noisy samples.}

\subsection{Ablation Studies}
\begin{figure}[t]
	\centering
	\includegraphics[width=\linewidth]{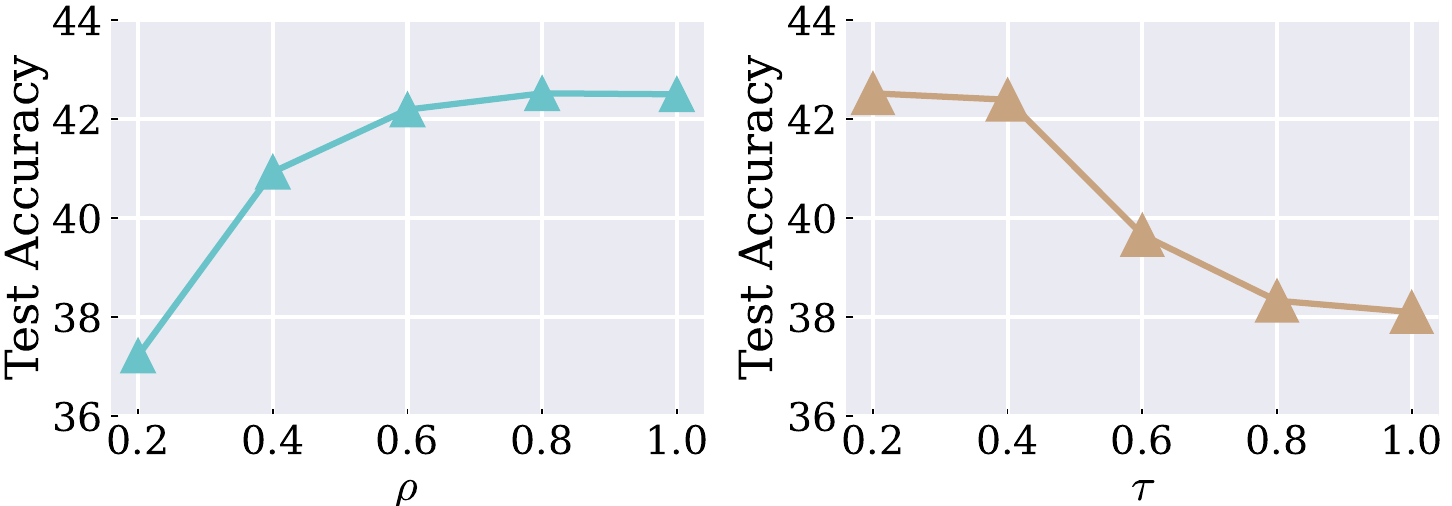}
    \vspace{-0.6cm}
	\caption{Hyper-parameter sensitivities of $\rho$ (left) and $\tau$ (right). Experiments are conducted on CIFAR100 (imbalance factor is 50 and noise rate is 20\%).}
	\label{figure5}
 \vspace{-0.3cm}
\end{figure}

{In this section, we study the influence of each proposed component (CBS, CSA, CR, and CM) and each hyper-parameter ($\rho$, $\tau$ and $\alpha$) in our method.}
We conduct ablation experiments on CIFAR10 and CIFAR100 with various imbalance factors (\ie, 1, 10, 50) and noise rates (\ie, 0\%, 20\%, 60\%).
The results are provided in Table~\ref{table:7} and Fig.~\ref{figure5}. 
Standard represents the conventional forward training using the cross-entropy loss.
CBS denotes the class-balance-based sample selection.
CSA indicates the confidence-based sample augmentation.
CR means consistency regularization.
ACM denotes the proposed average confidence margin.

\textbf{Effects of Class-Balance-based Sample Selection:}
Based on the aforementioned analysis, loss-based sample selection tends to cause learning bias in imbalanced datasets. 
In particular, tail class samples are prone to be identified as noise because under-learning leans to result in large loss values.
Our class-balance-based sample selection method ensures that the tail classes are not neglected during training. Accordingly, the learning bias issue is effectively alleviated.
As shown in Table~\ref{table:7}, employing CBS achieves notable performance gains compared to Standard in all experimental settings.

\textbf{Effects of Confidence-based Sample Augmentation:} 
The sample selection process may inevitably cause some noisy samples to be mistakenly selected into the clean subset. Thus, we propose confidence-based sample augmentation to enhance the reliability of selected clean samples.
Table~\ref{table:7} illustrates that adopting CSA boosts model performance, proving the effectiveness of CSA.
In particular, as the noise rate and imbalance factor increase, the performance gain is more significant.

\textbf{Effects of Consistency Regularization:}
Although we leverage EMA to re-assign labels for selected noisy samples, the corrected labels may still be inaccurate. The imperfect label correction is prone to causing performance degradation.
Accordingly, we propose consistency regularization to achieve enhancement in both feature extraction and model prediction. 
From Table~\ref{table:7}, we can find CR successfully boosts the model performance.
For example, employing CR yields a 3.95\% performance gain on CIFAR100 when noise rate and imbalance are 60\% and 10.

\textbf{Effects of Average Confidence Margin:}
In addition to the employment of consistency regularization, we introduce the average confidence margin to measure the confidence of corrected labels, aiming to promote model robustness when learning from label-corrected noisy samples.
Our training process uses a dynamic mechanism to continuously assess the impact of corrected labels on learning and generalization, rather than solely relying on model predictions at the current epoch. 
Accordingly, we discard label-corrected noisy samples with low confidence from training, alleviating their potential damage to the model.
As shown in Table~\ref{table:7}, employing ACM achieves considerable performance gains.

\textbf{Effects of Hyper-parameters:}
{We investigate the effects of hyper-parameters $\rho$, $\tau$ and $\alpha$ in our proposed method.}
We take the $\rho$ to control the proportion of selected clean samples per class.
{$\tau$ and $\alpha$ are set to control the number of reliable corrected labels and the loss weight in Eq.\ref{eq:13}, respectively.}
We provide the model performance under different $\rho$ and $\tau$ settings in Fig.~\ref{figure5}.
We can observe that a properly selected $\rho$ and $\tau$ can boost the model performance further.
When $\rho$ is 0.8 (\ie, 1-$\eta$), and $\tau$ is 0.2, our method achieves the highest performance on the test set on synthetic noisy CIFAR100, whose noise rate is 20\% and imbalance factor is 10.
{
Additionally, we further demonstrate the performance of our CBS under different loss weights $\alpha$ in Eq. \ref{eq:13}.
When the values of $\alpha$ are 0.5, 1.0, 1.5, and 2.0, the test accuracy is 41.37\%, 42.52\%, 40.41\%, and 39.81\%, respectively. 
This further validates the effectiveness of our loss function in Eq. \eqref{eq:13}.}

\section{Conclusion}
In this paper, we focused on the challenge of learning with noisy and imbalanced datasets. 
To address label noise and class imbalance simultaneously, we proposed a simple yet effective method based on balanced sample selection.
Our proposed method followed the semi-supervised learning paradigm and trained only one network in the training process.
Specifically, we proposed a class-balance-based sample selection strategy to divide samples into clean and noisy subsets in a class-balanced manner.
We then performed confidence-based sample augmentation to enhance the reliability of selected clean samples.
Afterward, we employed EMA to relabel selected noisy samples and filtered those with low confidence based on the average confidence margin metric.
Finally, consistency regularization was adopted on label-corrected noisy samples with high confidence to improve the robustness and stability of the model training.
Extensive experiments and ablation studies were conducted to substantiate the effectiveness and superiority of our proposed method.

\bibliographystyle{IEEEtran}
\bibliography{references}

\end{document}